# Uncertainty-Estimation with Normalized Logits for Out-of-Distribution Detection

Mouxiao Huang [a, b], Yu Qiao[*a]

[a]The Guangdong Provincial Key Laboratory of Computer Vision and Virtual Reality Technology, Shenzhen Institute of Advanced Technology, Chinese Academy of Sciences;
[b]University of Chinese Academy of Sciences
[*]Corresponding author: yu.qiao@siat.ac.cn

## ABSTRACT

Out-of-distribution (OOD) detection is critical for preventing deep learning models from making incorrect predictions to ensure the safety of artificial intelligence systems. Especially in safety-critical applications such as medical diagnosis and autonomous driving, the cost of incorrect decisions is usually unbearable. However, neural networks often suffer from the overconfidence issue, making high confidence for OOD data which are never seen during training process and may be irrelevant to training data, namely in-distribution (ID) data. Determining the reliability of the prediction is still a difficult and challenging task. In this work, we propose **U**ncertainty-**E**stimation with **N**ormalized **L**ogits (**UE-NL**), a robust learning method for OOD detection, which has three main benefits. (1) Neural networks with UE-NL treat every ID sample equally by predicting the uncertainty score of input data and the uncertainty is added into softmax function to adjust the learning strength of easy and hard samples during training phase, making the model learn robustly and accurately. (2) UE-NL enforces a constant vector norm on the logits to decouple the effect of the increasing output's norm from optimization process, which causes the overconfidence issue to some extent. (3) UE-NL provides a new metric, the magnitude of uncertainty score, to detect OOD data. Experiments demonstrate that UE-NL achieves top performance on common OOD benchmarks and is more robust to noisy ID data that may be misjudged as OOD data by other methods.

## 1. INTRODUCTION

As machine learning continues to advance and find applications in various fields, the importance of accurate model performance is increasingly evident. Unfortunately, many models fail to perform well in the real world because they were trained on data that closely resemble the training set, which can lead to poor generalization when applied to new, unseen data. This is a phenomenon known as OOD detection, and it is a critical challenge that must be addressed to improve the overall efficacy and reliability of machine learning models.

OOD detection is the task of identifying when a given input falls outside the distribution of the training data, which is also referred to as detecting novel or anomalous inputs[1]. When a model is presented with OOD inputs, its predictions become unreliable, which can lead to unexpected and potentially harmful consequences. For instance, an autonomous vehicle trained to recognize pedestrians using visual input from the training set may fail to detect a pedestrian that is not part of the training data, resulting in an accident. Similarly, a fraud detection system trained on historical data may fail to detect a new type of fraud that it has not seen before.

Therefore, OOD detection is an important challenge in machine learning that must be addressed to ensure that models can generalize to new, unseen data accurately. This is particularly important in domains where incorrect predictions can have severe consequences, such as healthcare, finance, and autonomous systems. By detecting OOD inputs, models can indicate when their predictions are unreliable and avoid making incorrect decisions.

In recent years, there has been a growing interest in developing techniques for OOD detection, and many researchers have proposed various methods for this task. The maximum softmax probability (MSP)[2] has been commonly used for OOD detection, based on the assumption that OOD inputs would result in lower confidence scores compared to ID inputs. However, deep neural networks can often produce overconfident predictions, even for inputs that are far from the training data, leading to doubts about using softmax confidence for OOD detection directly. As a result, many alternative OOD scoring functions have been proposed. ODIN[3] provides two strategies, temperature scaling and input preprocessing, for OOD detection. While, ODIN requires OOD data to tune hyperparameters, preventing it to generalize to other datasets. Energy-OOD[4] shows energy scores can be used to distinguish ID and OOD samples. LogitNorm[5] finds that neural networks trained through normalized logits mitigate the overconfidence issue and produces highly distinguishable confidence scores. Besides, some other related works can achieve OOD detection. Methods like PFE[6], DUL[7], RTS[8] are used to detect OOD inputs with uncertainty score trained by modeling embeddings as a Guassian distribution. MagFace[9] and AdaFace[10] use magnitudes of face embeddings to present face quality, which reveals the distribution of inputs to some extent and can also be applied as an OOD detection metric[8].

In this work, we propose Uncertainty-Estimation with Normalized Logits (UE-NL), an efficient method to learn a reliable classification and OOD detection model. Motivated by DUL and RTS, we take a probabilistic view of the classification method using softmax with temperature scaling and find that the temperature scalar in softmax quantifies the input uncertainty. UE-NL acts like a Bayesian network[11,12,13], regularizes confidence and learns the embeddings and uncertainty scores simultaneously. It adjusts the learning strength of easy and hard samples during training phase, making the learning process robustly. And the learnable uncertainty is proved to be a good metric for OOD detection. Moreover, we normalize the logits output from the last fully connected layer of prediction network to decouple the effect of the increasing output's norm from optimization process, mitigating the overconfidence issue which classification models usually suffer from. UE-NL requires no OOD data to tune its hyperparameters and adds only negligible computation cost to the original classification model.

## 2. METHODOLOGY

### 2.1 Preliminaries

We are dealing with a multi-class classification problem that involves supervised learning. The input space is denoted as $X$ and the label space as $Y$, which consists of $k$ classes represented by integers from 1 to $k$. The training dataset, $D = \{x_i, y_i\}_{i=1}^{N}$, consists of $N$ data points that are independently and identically sampled from a joint data distribution. The marginal distribution over $X$, which represents the in-distribution, is denoted as $Dist_{ID}$. Using the training dataset, we aim to learn a classifier, $f$, with trainable parameter $\theta$ that maps an input to the output space. Moreover, cross-entropy loss (CELoss) with softmax function is commonly used in classification models optimization, shown as below:

$$CELoss(f(x;\theta), y) = -\log \frac{e^{f_y(x;\theta)}}{\sum_i e^{f_i(x;\theta)}} \tag{1}$$

The OOD detection task is a binary-classification task. Ideally, the test data should follow a distribution $Dist_{OOD}$ that does not overlap with $Dist_{ID}$. Thus, the OOD detection can be formulated as:

$$x_i \text{ is } \begin{cases} ID, \text{ if } S(x_i) \geq \beta \\ OOD, \text{ if } S(x_i) < \beta \end{cases} \tag{2}$$

Where $S(x)$ is a scoring function that determines a sample as ID or OOD. $\beta$ is the threshold for the decision.

### 2.2 Uncertainty-Estimation with Normalized Logits

In the following, we elaborate the details of the proposed method Uncertainty-Estimation with Normalized Logits (UE-NL). The algorithm is described in Table 1. Specifically, $f$ represents the classification backbone that outputs logits $p$ and embedding $e$. And $g$ is the uncertainty prediction module with embedding $e$ as its input. $g$ has one linear layer and one batch normalization layer and follows an exponential function, which can be formulated as $g(e) = exp(BN(w_g \cdot e + b_g))$. Although the sampling operation is non-differentiable, which hinders the flow of gradient backpropagation during model training, we can use re-parameterization[7,8] trick to enable the model to compute gradients as usual. We sample a random noise value $\varepsilon$ from a normal distribution that is independent of the model's parameters and compute the resampled uncertainty score.

Table 1. The algorithm of the proposed method: UE-NL

---
**Algorithm: UE-NL**

**Input:** Image and label $(x, y)$, dimension of resampled uncertainty score $\delta$, KL weight $\lambda$

**Parameter:** Backbone $\theta$, uncertainty net $\varphi$

**Output:** Logits vector $p$, embedding $e$, uncertainty score $u$

Predict the logits vector and embedding:
$$p, e = f(x; \theta)$$
Predict the uncertainty score:
$$u = g(e; \varphi) \in \mathbb{R}$$
Resample the uncertainty score and normalize the logits:
$$\hat{u} = \sum_i u_i \cdot \varepsilon_i^2, \; \bar{p} = \frac{p}{\|p\|}; \; \varepsilon \sim \mathcal{N}(0, I) \in \mathbb{R}^\delta$$

---

In the proposed method, UE-NL, we use CELoss with predicted uncertainty score and KLLoss to train the model, which is given by:

$$Loss = CELoss + \lambda \cdot KLLoss \begin{cases} CELoss = -\log \frac{e^{\frac{\bar{p}}{\hat{u}}}}{\sum_i e^{\frac{\bar{p}}{\hat{u}}}} \\ KLLoss = \sum_{i=1}^{\delta} KL(\mathcal{N}(0, u), \mathcal{N}(0, I)) \end{cases} \quad (1)$$

Where $\lambda$ is the weight of KLLoss, a hyperparameter that balances the two losses.

## 3. EXPERIMENTS

### 3.1 Experiments setup

**Datasets.** Follow related prior works, we use CIFAR-10[14] and CIFAR-100[14] as our training data for experiments. They are commonly used in OOD detection task. In our implementation, all images are resized to $32 \times 32$ and we use standard splits, training data and test data has 50,000 images and 10,000 images, respectively. For OOD detection performance evaluation, we use four benchmarks as OOD data, Textures[15], SVHN[16], LSUN-C[17], LSUN-R[17] and iSUN[18]. Details can be seen in Table 2.

Table 2. Datasets information.

| Dataset | Number of Images | Instruction |
| --- | --- | --- |
| CIFAR-10 / CIFAR-100 | Train: 50,000 / Test: 10,000 | General objects images |
| Textures | 5,640 | Describable textual images |
| SVHN | 26,032 | House numbers images |
| LSUN-C | 10,000 | Scene understanding dataset |
| LSUN-R | 10,000 | Scene understanding dataset |
| iSUN | 8,925 | Natural scene images |

**Evaluation metrics.** There are three main metrics to evaluate the performance of OOD detection: (1) FPR95 is the false positive rate when the true positive rate is 95%; (2) AUC is the area under the receiver operating characteristic curve, a performance metric that measures the ability of a binary classification model to distinguish between positive and negative classes. (3) AUPR is the area under the precision-call curve and is often used when the positive class is rare or when the focus is on correctly identifying positive examples.

**Implementation details.** Following the settings of state-of-the-art (SOTA) work LogitNorm, we use WRN-40-2[19] as our backbone and train classification models for 200 epochs using Stochastic Gradient Descent (SGD) with a momentum of 0.9, a weight decay of 0.0005, a dropout rate of 0.3, and a batch size of 128. The initial learning rate is set to 0.1, and it is decreased by a factor of 10 at the 80th and 140th epochs. We use Gaussian noises as validation dataset to tune hyperparameters and finds the best model according to validation performance. And then we conduct experiments on test dataset to report every evaluation metric score and average score. For a fair comparison, all methods are reproduced with default settings mentioned above. And other hyperparameters in each method are derived from their original papers. In particular, we use temperature scalar $T$=1000 and $\varepsilon$=0.0014 in ODIN. For Energy-OOD, we use $T = 0.1$ in CIFAR-10 and $T = 0.01$ in CIFAR-100. For LogitNorm, temperature parameter is set as $T = 0.04$. Besides methods based on general image classification, we also reproduce SOTA uncertainty-based for OOD detection method, RTS, whose hyperparameters, $\delta$ and $\lambda$ are tuned to achieve its best performance.

## 3.2 Results

**OOD detection performance comparison.** To evaluate the OOD detection performance of UE-NL, we use the most commonly used metrics FPR95 / AUC / AUPR (%, best score is indicated in bold) and conduct experiments on CIFAR-10 and CIFAR-100. Table 3 and Table 4 report scores of different methods when detecting various OOD test dataset. As we can see in Table 3, UE-NL achieves the best in most test datasets. In CIFAR-100, our method also improves the performance. And the mean results of five datasets further demonstrate that UE-NL outperforms by a meaningful margin than other methods. Especially in CIFAR-100, compared with SOTA, our method improves in all indicators, with an average increase of 9.0 / 4.5 / 3.5 (%), respectively.

Table 3. OOD performance comparison trained on CIFAR-10. Evaluation metrics are FPR95 / AUC / AUPR (%), respectively.

| OOD Method | Textures | SVHN | LSUN-C | LSUN-R | iSUN | Mean |
|---|---|---|---|---|---|---|
| CE | 61.1 / 76.2 / 76.6 | 59.9 / 86.4 / 65.8 | 43.2 / 83.7 / 72.2 | 57.9 / 81.2 / 74.8 | 62.9 / 77.6 / 73.5 | 57.0 / 81.0 / 72.6 |
| ODIN | 74.5 / 55.5 / 63.1 | 72.3 / 66.3 / 36.6 | 63.3 / 56.1 / 48.0 | 57.4 / 72.8 / 65.2 | 64.1 / 69.1 / 65.3 | 66.3 / 64.0 / 55.6 |
| Energy-OOD | 57.8 / 75.0 / 75.4 | 55.3 / 85.5 / 62.6 | 39.0 / 83.0 / 70.9 | 52.0 / 81.5 / 74.9 | 57.9 / 77.9 / 73.5 | 52.4 / 80.6 / 71.5 |
| LogitNorm | 40.7 / **92.2** / **95.5** | 23.3 / **96.2** / 93.6 | **3.8** / 99.1 / 99.2 | **15.6** / 97.2 / 97.7 | 19.8 / 96.7 / 97.5 | 20.6 / 96.3 / 96.7 |
| RTS | 53.3 / 88.9 / 83.7 | **10.9** / **97.0** / **98.6** | 7.2 / 98.4 / 98.4 | 44.4 / 89.5 / 88.6 | 62.9 / 86.3 / 84.4 | 35.7 / 92.0 / 90.8 |
| UE-NL | **39.1** / 92.0 / 95.1 | 22.6 / 96.2 / 93.7 | 3.9 / **99.2** / **99.3** | **15.6** / **97.5** / **97.8** | **18.4** / **97.1** / **97.9** | **19.9** / **96.4** / **96.8** |

Table 4. OOD performance comparison trained on CIFAR-100. Evaluation metrics are FPR95 / AUC / AUPR (%), respectively

| OOD Method | Textures | SVHN | LSUN-C | LSUN-R | iSUN | Mean |
|---|---|---|---|---|---|---|
| CE | 77.9 / 68.3 / 72.5 | 80.8 / 60.9 / 31.1 | 71.9 / 72.6 / 64.5 | 73.7 / 68.9 / 62.9 | 75.4 / 69.4 / 66.9 | 75.9 / 68.0 / 59.6 |
| ODIN | 79.6 / 60.9 / 68.1 | 89.9 / 49.6 / 27.8 | 79.5 / 55.0 / 49.1 | **68.2** / 68.5 / 61.1 | 67.7 / 70.0 / 65.9 | 76.9 / 60.8 / 54.4 |
| Energy-OOD | 75.8 / 70.6 / 74.2 | 78.6 / 63.7 / 32.7 | 69.8 / 73.1 / 63.9 | 70.9 / 71.5 / 64.4 | 72.2 / 72.1 / 68.4 | 73.5 / 70.2 / 60.8 |
| LogitNorm | 80.3 / 76.5 / 84.8 | 65.3 / 88.3 / 83.3 | 27.1 / 94.9 / 95.4 | 89.1 / 67.4 / 69.2 | 88.6 / 68.2 / 71.9 | 70.1 / 79.1 / 80.9 |
| RTS | 80.1 / 77.1 / 85.4 | **63.6** / **88.8** / 83.7 | 37.6 / 92.8 / 93.6 | 80.4 / 73.3 / 74.7 | 83.8 / 72.3 / 75.8 | 69.1 / 80.8 / 82.6 |
| UE-NL | **72.1** / **80.4** / **86.5** | 67.6 / 87.4 / 81.7 | **20.8** / **96.3** / **96.6** | 68.7 / **81.3** / **82.2** | **71.3** / **80.9** / **83.6** | **60.1** / **85.3** / **86.1** |

**Distribution of scores of OOD metric.** Following LogitNorm, we visualize the distribution of OOD detection metric scores of CE, LogitNorm and UE-NL on ID test data and OOD data. While CE and LogitNorm use MSP as their metric, our method uses an uncertainty score instead. LogitNorm claims to have better distinguishability between ID and OOD data, but when tested on a more challenging task such as CIFAR-100, its performance drops significantly. Figure 1 shows that the distributions for CE and LogitNorm are more overlapped, whereas UE-NL has a less overlapping distribution, indicating that our method can more effectively differentiate between OOD and ID data.

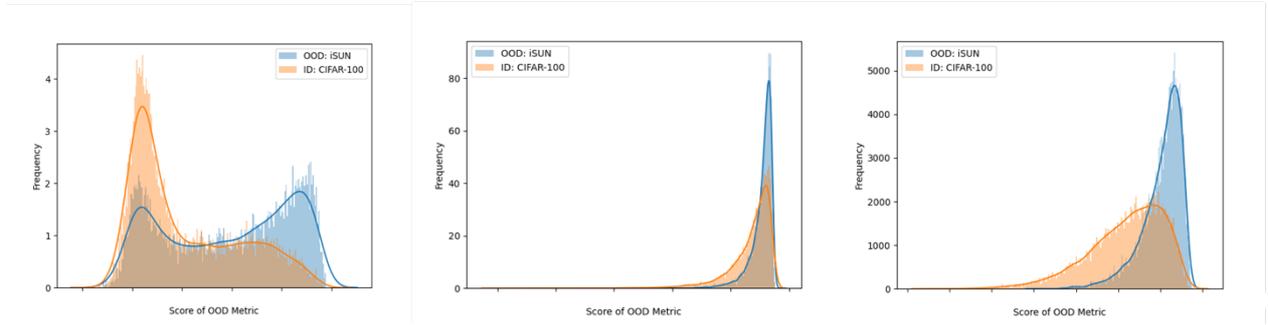

Figure 1. Distribution of scores of OOD metric: CE, LogitNorm and UE-NL, respectively. Training data: CIFAR-100.

**Classification accuracy in ID data.** The main purpose of a classifier is to make correct decisions, which requires high accuracy in identifying data. LogitNorm claims to maintain classification accuracy. Our method UE-NL not only achieves the best performance in the OOD detection task, but also maintains and improves classification accuracy. The learnable uncertainty score in the training phase plays the role of an adaptive temperature scalar that can adjust the learning strength of easy and hard samples, which helps with optimization.

Table 5. Error rate and accuracy in ID data classification of different methods.

| Metric / Method | Error Rate / Acc (CIFAR-10) | Error Rate / Acc (CIFAR-100) |
|---|---|---|
| CE | 5.38 / 94.62 | 25.70 / 74.30 |
| LogitNorm | 5.65 / 94.35 | 24.74 / 75.26 |
| RTS | 5.35 / 94.65 | **24.34 / 75.66** |
| UE-NL | **5.33 / 94.67** | 24.54 / 75.46 |

## 4. ABLATION STUDY

Following RTS, we also conduct experiments on CIFAR-10 in this part to study the effect of hyperparameters.

**Shape of resampled uncertainty score.** The hyperparameter $\delta$ represents the dimension of uncertainty score after re-parameterization. The results are shown in Table 6 and it shows that UE-NL achieves the best performance when $\delta = 32$.

Table 6. OOD detection and classification performance with different $\delta$ ($\lambda = 0.1$).

| $\delta$ | FPR95 / AUC / AUPR (%, mean) | Error Rate / Acc |
|---|---|---|
| 16 | 36.4 / 92.9 / 92.6 | 5.7 / 94.3 |
| 32 | **19.9 / 96.4 / 96.8** | **5.3 / 94.7** |
| 64 | 22.8 / 95.8 / 95.9 | 5.6 / 94.4 |

**Weight of KL loss.** The KL divergence loss serves as a regularization term, preventing the uncertainty scale from becoming unbounded. Through results shown in Table 7, we set the weight $\lambda$ as 0.1.

Table 7. OOD detection and classification performance with different $\lambda$ ($\delta$ = 32).

| $\lambda$ | FPR95 / AUC / AUPR (%, mean) | Error Rate / Acc |
|---|---|---|
| 0.01 | 30.0 / 94.4 / 95.5 | 5.5 / 94.5 |
| 0.1 | **19.9 / 96.4 / 96.8** | **5.3 / 94.7** |
| 1.0 | 30.0 / 93.6 / 94.2 | 5.6 / 94.4 |

## 5. CONCLUSION

In this paper, we propose Uncertainty-Estimation with Normalized Logits (UE-NL), a simple and effective robust learning method for OOD detection. Classification models incorporating UE-NL predict a learnable uncertainty score during training, allowing the network to adjust the learning strength of easy and hard samples, resulting in more stable optimization. Furthermore, normalization of logits can decouple the effect of increasing output norm during the training process and mitigate overconfidence issues. Extensive experiments demonstrate that UE-NL not only provides a new metric that achieves state-of-the-art (SOTA) performance for OOD detection in general classification tasks but is also beneficial for classifying In-Distribution (ID) data. The method is easy to implement in any classification model and adds only negligible computation cost, without requiring extra OOD data during training. We believe that UE-NL can inspire more researchers to advance the OOD detection field.

## REFERENCES

[1] Yang, Jingkang, et al. "Generalized out-of-distribution detection: A survey." arXiv preprint arXiv:2110.11334 (2021).
[2] Hendrycks, Dan, and Kevin Gimpel. "A baseline for detecting misclassified and out-of-distribution examples in neural networks." arXiv preprint arXiv:1610.02136 (2016).
[3] Liang, Shiyu, Yixuan Li, and Rayadurgam Srikant. "Enhancing the reliability of out-of-distribution image detection in neural networks." arXiv preprint arXiv:1706.02690 (2017).
[4] Liu, Weitang, et al. "Energy-based out-of-distribution detection." Advances in neural information processing systems 33: 21464-21475 (2020).
[5] Wei, Hongxin, et al. "Mitigating neural network overconfidence with logit normalization." International Conference on Machine Learning. PMLR, 2022.
[6] Shi, Yichun, and Anil K. Jain. "Probabilistic face embeddings." Proceedings of the IEEE/CVF International Conference on Computer Vision. 2019.
[7] Chang, Jie, et al. "Data uncertainty learning in face recognition." Proceedings of the IEEE/CVF conference on computer vision and pattern recognition. 2020.
[8] Shang, Lei, et al. "Improving Training and Inference of Face Recognition Models via Random Temperature Scaling." arXiv preprint arXiv:2212.01015 (2022).
[9] Meng, Qiang, et al. "Magface: A universal representation for face recognition and quality assessment." Proceedings of the IEEE/CVF Conference on Computer Vision and Pattern Recognition. 2021.
[10] Kim, Minchul, Anil K. Jain, and Xiaoming Liu. "Adaface: Quality adaptive margin for face recognition." Proceedings of the IEEE/CVF Conference on Computer Vision and Pattern Recognition. 2022.
[11] Gal, Yarin, and Zoubin Ghahramani. "Dropout as a bayesian approximation: Representing model uncertainty in deep learning." international conference on machine learning. PMLR, 2016.
[12] Theobald, Claire, et al. "A Bayesian Neural Network based on Dropout Regulation." arXiv preprint arXiv:2102.01968 (2021).
[13] Guan, Weipeng, et al. "High-speed robust dynamic positioning and tracking method based on visual visible light communication using optical flow detection and Bayesian forecast." IEEE Photonics Journal 10.3 (2018)
[14] Krizhevsky, Alex, and Geoffrey Hinton. "Learning multiple layers of features from tiny images." (2009): 7.
[15] Cimpoi, M., Maji, S., Kokkinos, I., Mohamed, S., and Vedaldi, A. Describing textures in the wild. In Proceedings of the IEEE Conference on Computer Vision and Pattern Recognition, pp. 3606–3613, 2014.

[16] Netzer, Y., Wang, T., Coates, A., Bissacco, A., Wu, B., and Ng, A. Y. Reading digits in natural images with unsupervised feature learning. NIPS Workshop on Deep Learning and Unsupervised Feature Learning, 2011.

[17] Yu, F., Seff, A., Zhang, Y., Song, S., Funkhouser, T., and Xiao, J. Lsun: Construction of a large-scale image dataset using deep learning with humans in the loop. arXiv preprint arXiv:1506.03365, 2015.

[18] Xu, P., Ehinger, K. A., Zhang, Y., Finkelstein, A., Kulka- rni, S. R., and Xiao, J. Turkergaze: Crowdsourcing saliency with webcam based eye tracking. arXiv preprint arXiv:1504.06755, 2015.

[19] Zagoruyko, S. and Komodakis, N. Wide residual networks. arXiv preprint arXiv:1605.07146, 2016.